\documentclass{article}

\usepackage{microtype}

\usepackage{graphicx}
\usepackage{booktabs} % for professional tables

\usepackage{algorithm}
 \usepackage[noend]{algorithmic}
\renewcommand{\algorithmiccomment}[1]{ $\hfill\triangleright$ #1}

\usepackage[utf8]{inputenc} % allow utf-8 input
\usepackage[T1]{fontenc}    % use 8-bit T1 fonts
\usepackage{url}            % simple URL typesetting
\usepackage{booktabs}       % professional-quality tables
\usepackage{amsfonts}       % blackboard math symbols
\usepackage{nicefrac}       % compact symbols for 1/2, etc.
\usepackage[normalem]{ulem}
\usepackage{enumitem}
\usepackage{rotating}
\usepackage{balance}
\usepackage{amsmath}
\usepackage{amssymb}
\usepackage{txfonts}
\usepackage{multirow}
\usepackage{xcolor}
\usepackage{soul}
\usepackage{tikz}
\usepackage{float}
\usepackage{wrapfig}
\usepackage{caption}
\usepackage{subfig}
\usepackage{multirow}

\usepackage{hyperref}
\usepackage{cleveref}
\crefformat{section}{\S#2#1#3} % see manual of cleveref, section 8.2.1
\crefformat{subsection}{\S#2#1#3}
\crefformat{subsubsection}{\S#2#1#3}

\setlist{noitemsep, leftmargin=*, topsep=0pt, partopsep=0pt}
\newcommand{\SYSTEM}{\textsc{Nurd}}
\newcommand{\slow}{stra}
\newcommand{\fast}{run}

\usepackage[accepted]{mlsys2022}

\mlsystitlerunning{\SYSTEM{}: Negative-Unlabeled Learning for Online Datacenter Straggler Prediction}

\begin{document}

\twocolumn[
\mlsystitle{\SYSTEM{}: Negative-Unlabeled Learning for Online Datacenter Straggler Prediction}

% unless you've provided the [accepted] option to the mlsys2022


\mlsyssetsymbol{equal}{*}

\begin{mlsysauthorlist}
\mlsysauthor{Yi Ding}{to}
\mlsysauthor{Avinash Rao}{goo}
\mlsysauthor{Hyebin Song}{ed}
\mlsysauthor{Rebecca Willett}{goo}
\mlsysauthor{Henry Hoffmann}{goo}
\end{mlsysauthorlist}

\mlsysaffiliation{to}{MIT CSAIL}
\mlsysaffiliation{ed}{Pennsylvania State University}
\mlsysaffiliation{goo}{University of Chicago}

\mlsyscorrespondingauthor{Yi Ding}{ding1@csail.mit.edu}

\mlsyskeywords{Machine Learning, MLSys}

\vskip 0.3in

\begin{abstract}
Datacenters execute large computational jobs, which are composed of smaller tasks. A job completes when all its tasks finish, so \emph{stragglers}---rare, yet extremely slow tasks---are a major impediment to datacenter performance.  Accurately predicting stragglers would enable proactive intervention, allowing datacenter operators to mitigate stragglers before they delay a job. While much prior work applies machine learning to predict computer system performance, these approaches rely on \emph{complete labels}---i.e., sufficient examples of all possible behaviors, including straggling and non-straggling---or \emph{strong assumptions about the underlying latency distributions}---e.g., whether Gaussian or not. Within a running job, however, none of this information is available until stragglers have revealed themselves when they have already delayed the job. To predict stragglers accurately and early without labeled positive examples or assumptions on latency distributions,  this paper presents \emph{\SYSTEM{}}, a novel \textit{\textbf{N}}egative-\textit{\textbf{U}}nlabeled learning approach with \textit{\textbf{R}}eweighting and \textit{\textbf{D}}istribution-compensation that only trains on negative and unlabeled streaming data. The key idea is to train a predictor using finished tasks of non-stragglers to predict latency for unlabeled running tasks, and then reweight each unlabeled task's prediction based on a weighting function of its feature space. We evaluate \SYSTEM{} on two production traces from Google and Alibaba, and find that compared to the best baseline approach, \SYSTEM{} produces 2--11 percentage point increases in the F1 score in terms of prediction accuracy, and 2.0--8.8 percentage point improvements in job completion time.

\end{abstract}

]

\printAffiliationsAndNotice{}  % leave blank if no need to mention equal contribution
%\printAffiliationsAndNotice{\mlsysEqualContribution} % otherwise use the standard text.

\section{Introduction}\label{sec:intro}

Stragglers impede job completion in datacenter-scale computing. Here, a computational \emph{job} is split into many \emph{tasks}, each of which is executed in parallel on different machines before their results are aggregated when the last task completes. \emph{Stragglers} are rare, extremely slow tasks within a job that can degrade overall performance---by as much as 30--50\%~\cite{ananthanarayanan2013effective,reiss2011google,ZhengL18}. A straggler is commonly defined as a task with at least 90th percentile (p90) latency; i.e., at least 90\% of tasks finish earlier than the straggler~\cite{hao2017mittos,hao2020linnos}. We refer to stragglers as the \emph{positive} class since they are the minority and abnormal, and non-stragglers are the \emph{negative} class since they are the majority and expected~\cite{chandola2009anomaly}.

Mitigating stragglers is a fundamental problem in datacenters~\cite{zhou2021comprehensive,belay2014ix,Adya2016slicer,Handley2017re,Haque2015few,nelson2015latency,ayers2019asmdb}. Recent work uses predictive models to monitor executing tasks and predict stragglers before they reveal themselves with long run times~\cite{ananthanarayanan2010reining,ren2015hopper,yadwadkar2014wrangler,zhou2020falcon}. Once a straggler is correctly predicted, proactive interventions---such as relaunching the same task on a different machine---will be triggered to mitigate the straggling behavior~\cite{ananthanarayanan2013effective,aktas2017effective,aktacs2019straggler}. Machine learning techniques have been applied to model the complicated, nonlinear relationships between features (e.g., CPU utilization) and computation behavior (e.g., latency)---a recent survey has details~\cite{penney2019survey}. However, most existing work either heavily relies on \emph{complete labels}---i.e., observing labeled samples from all classes at training---or \emph{strong assumptions about the underlying latency distribution}---e.g., whether Gaussian or not. When predicting stragglers on live data---i.e., running jobs in the datacenter---stragglers are not revealed early because they finish last. Therefore, there are insufficient labels in the training set due to a lack of positive examples of stragglers, and it is hard to pre-specify the latency distribution for each job, which render most learning methods ineffective for straggler prediction within a running job. Moreover, since the characteristics of each job are usually unique in datacenters~\cite{reiss2012heterogeneity,guo2019}, it is difficult to train a model on one job and apply it to another directly. Therefore, this paper proposes a technique that constructs a unique predictive model for each job on the fly---that is, as the job is running and before stragglers reveal themselves with long completion time.

We present \SYSTEM{}, a novel negative-unlabeled learning approach for online straggler prediction that requires no labeled positive examples or assumptions on the latency distributions. \SYSTEM{} uses finished tasks (i.e., negative examples, non-stragglers) to train a model to predict latency as a function of observed task features for running tasks. \SYSTEM{}'s key insight is that this predictor will be biased towards non-stragglers, so  it then reweights these latency predictions using a function of task features---i.e., each running task's probability of being included in the set of finished tasks given its observed features. Intuitively, this weighting function indicates how dissimilar a particular running task's features are from those that are finished; i.e., it preserves latency predictions for tasks that are similar to finished tasks (i.e., non-stragglers), and increases predicted latency for those that are different. With this reweighting scheme, \SYSTEM{} predicts stragglers early and accurately by reducing the prediction bias due to a lack of stragglers at training.

To summarize, our main contributions are as follows:
\begin{itemize}
\item We propose a novel negative-unlabeled learning approach based on reweighting predictions and demonstrate its efficacy for online straggler prediction when no labeled stragglers exist in the training set.
\item We evaluate \SYSTEM{} on Google production traces~\cite{reiss2011google}, where we observe an 11 percentage point increase in the F1 score and 3.8--8.8 percentage point improvement in job completion time relative to the best baseline approach. Similarly, on Alibaba production traces~\cite{alibaba}, we see a 2 point increase in the F1 score and an 2.0--3.5 point improvement in job completion time.
\item We release the code in \url{https://github.com/y-ding/nurd-mlsys22-code}.
\end{itemize}

Stragglers significantly hamper system performance in modern datacenters. By identifying stragglers accurately and early for running jobs, \SYSTEM{} provides a novel online learning approach that does not require labeled positive examples of stragglers or assumptions on latency distributions. This work offers a new direction in which systems community can apply machine learning techniques that can generalize without heavy reliance on carefully curating training sets.

\section{Background}\label{sec:backgorund}

\textbf{Datacenter terminology.} Datacenter-scale computations, or \emph{jobs}, are composed of sub-computations called \emph{tasks}. Because datacenter performance is critical, jobs are continually monitored and tasks' behavior in a variety of metrics are recorded at regular time \emph{checkpoints}~\cite{reiss2011google,reiss2012heterogeneity}. These recorded measurements are a set of \emph{features} that characterize each task. The choice of metrics to monitor (and thus features to capture) depends on the specific system on which the job is deployed. Ideally, datacenters would record metrics related to resource usage, microarchitectural behavior, and job scheduling~\cite{ZhengL18}.  In such scenario, a straggler prediction algorithm could be applied at each checkpoint using the features from each task to predict its future latency. If a particular task is expected to straggle---i.e., exceed an operator-specified latency threshold---then the job scheduler or human operator could be alerted to trigger intervention to mitigate stragglers.

\textbf{Straggler mitigation.} Straggler mitigation is an important part of datacenter scheduling~\cite{dean2013,Schwarzkopf2018research,aktacs2019straggler,dean2004mapreduce,zaharia2008improving,ananthanarayana2011scarlett}. Performance-aware schedulers predict which tasks are likely to straggle and then allocate additional resources to them~\cite{ananthanarayanan2010reining,yadwadkar2014wrangler,ren2015hopper}. Wrangler is a typical system like this, using linear support vector machines to classify stragglers by oversampling stragglers to deal with imbalanced labels in the training set~\cite{yadwadkar2014wrangler}. Another example is LinnOS, which uses neural networks to predict anomalous I/O latency by training on tens of thousands of I/O operations from a single hardware device with known latency~\cite{hao2020linnos}. A clear limitation of these approaches is that they require positive examples of stragglers. To the best of our knowledge, prior performance-aware schedulers assume that they have access to at least some examples of stragglers to train a model.  This is a strong assumption that does not hold if users develop new jobs that are different from existing jobs (which is the common case for datacenter jobs~\cite{reiss2012heterogeneity,guo2019}) or if datacenters install new hardware that induces new causes of straggling behavior. Thus, there is a need for performance-aware scheduling approaches that produce accurate predictions without positive examples of stragglers.

\textbf{Problem formulation.} Given the above discussion, we formalize the online straggler prediction problem as follows. Imagine we have $T$ time checkpoints. At the $t$-th checkpoint where $t\in[T]$, $n$ tasks are observed for a job, and the $i$-th task is associated with a feature vector $x_{ti}\in\mathbb{R}^d$, where $d$ is the number of features (measurements) that characterize the task.  Task $i$ has true latency $y_i \in\mathbb{R}_+$. Let $\tau^{\rm \slow}$ denote the target latency threshold that denotes straggling (e.g., the p90 latency), and $S := \{i \in [n]: y_i \geq \tau^{\rm \slow}\}$ denote the set of tasks that are true stragglers. Our goal is to identify the straggler set $S$. The challenge is that we do not observe $y_i$ for all tasks at the $t$-th checkpoint. Rather, we only observe $y_i$ when $y_i \leq \tau_{t}^{\rm \fast} \leq \tau^{\rm \slow}$, where $\tau_{t}^{\rm \fast}$ is the latency at $t$-th time checkpoint. Let $F_t:= \{i \in [n]: y_i \leq \tau_t^{\rm \fast}\}$ denote the tasks that finish before $t$-th time checkpoint, and $R_t$ be the list of tasks that are still running at the $t$-th time checkpoint. At each $t$-th checkpoint, given $x_{ti}$ for $i\in[n]$ and $y_i$ for $i \in F_t$, we estimate a set of stragglers $\hat S$ from the unfinished tasks at $\tau_t^{\rm \fast}$. Our goal is to correctly identify stragglers, so that intervention can occur as early as possible.

\section{Related Work}
\label{sec:related}

%\hh{I really like this summary.  I think we could make it even stronger by adding a very short couple of sentences that connect the formalism in the previous paragraph (prior section) to the abstract concepts in this summary, and then connect all that to the specfici problem of straggler detection.  So, something like the following: Given the above, we have a set of tasks in a job, which can be characterized by their features $x_{ti}$.  These features include behavior that is observable for all running tasks (such as IPC, DRAM bandwidth, cache miss rates, etc).  The goal is to detect stragglers, or tasks whose latency is especially long. NOTE: I wrote that and thought about it for a bit and now I am not sure I agree with my own advice.  I note that "response" is confusing in the below and we never defined it.  I think that is too abstract and if we change response space to latency in the second bullet and then change "pre-specified data" to "known latency" distribution that might be all we need to map the abstract concepts to the specific problem of latency prediction. }

We notice the following limitations from the existing approaches applied to online straggler prediction:
\begin{itemize}
    \item Difficulty of accounting for the drift between training and test distribution (supervised learning in~\cref{sec:supervised}).
    \item Only using information from the feature space and ignoring the observed latency (outlier detection in~\cref{sec:outlier}).
    \item Incorrect independence assumptions on labels given features (PU learning in~\cref{sec:pu}).
    \item Heavy reliance on pre-specified latency distribution (censored and survival regression in~\cref{sec:survival}).
\end{itemize}

\subsection{Supervised Learning}\label{sec:supervised}

Supervised learning uses labeled samples to learn a predictor $h_t$ so that $\hat y_{ti} = h_t(x_{ti})$ for $i$-th task at $t$-th checkpoint; this predictor could estimate the latency of the unlabeled samples. Critically, however, the distribution of the unlabeled samples is different from that of the labeled samples; that is, the nature of the straggler prediction problem necessitates a distribution drift between training and prediction, and we must be robust to that drift. Without accounting for this drift, latency predictions for stragglers will be heavily biased~\cite{quinonero2009dataset,zhang2013domain}.

\subsection{Outlier Detection (Unsupervised Learning)}\label{sec:outlier}

Outlier---or anomaly---detection is a family of unsupervised learning techniques that identifies rare events which differ from the general distribution of a population~\cite{chandola2007outlier,alam2019zero,sipple2020interpretable}. These techniques separate nominal and anomalous distributions based on the observations in the feature space only. Although stragglers can be thought of as outliers, our online straggler prediction problem is critically different from outlier detection problems studied in the literature because stragglers---by definition---are outliers in latency, which are not necessarily outliers in the feature space. As such, while we have access to each task's features at $t$-th time checkpoint, their latency values are revealed only up to time $t$.  
% Strategies of identifying stragglers only based on the separation of observations in feature space tend to have limited discriminative power (see Table~\ref{tbl:pred-sum} in \cref{sec:evaluation}), and identifying stragglers based on task latencies is impossible because the full latency distribution is not available at any training time point. This is in sharp contrast to standard settings in outlier detection problems, where at any timepoint $t$, a random sample is obtained from a mixture of nominal and anomalous distributions. 
To address the issue of only using information from feature space, \SYSTEM{} proposes to reweight the predicted latency with a weighting function to  reduce the prediction bias in latency due to a lack of stragglers at training. Empirically, we evaluate fourteen existing outlier detection methods in \cref{sec:evaluation} to demonstrate that outlier detection methods have limited discriminative power in identifying stragglers within running jobs.

\subsection{PU Learning (Semi-supervised Learning)}\label{sec:pu}

Positive-unlabeled (PU) learning is a family of semi-supervised learning techniques that use both labeled and unlabeled samples to train a classifier~\cite{bekker2020learning}. PU learning approaches learn from the two sets of examples, where the first set (positive) only contains examples from the first class, while the other set (unlabeled) contains examples from both classes. Existing PU learning~\cite{lee2003learning,elkan2008learning,mordelet2014bagging,kiryo2017positive} assumes that observations of the labels are independent of the features given the classes (positive or negative); that is, the labeled examples are a random sample from the positive examples. However, this assumption is violated in online straggler prediction because only some non-stragglers with lower latency values have a chance to be sampled, while other non-stragglers with higher latency values are not included in the labeled set.

\subsection{Censored and Survival Regression}\label{sec:survival}

Censored regression is a family of techniques to handle the situation where the value to be predicted (latency in this case) is censored; i.e., some values are missing but known to exceed certain thresholds~\cite{powell1986censored}. Survival regression is a related field that predicts when a system will survive beyond a certain time point.  The latency variable in the online straggler prediction problem can be viewed as being censored at each checkpoint $t$ because the latency values above $t$ are not revealed.
There are both linear (Tobit~\cite{tobin1958estimation}) and non-linear (Grabit \cite{sigrist2019grabit}) methods for censored regression.  The Cox proportional hazard (CoxPH) model is a popular approach for survival analysis  \cite{Cox1972-be}.   All three methods can be used to estimate latency with incomplete labels: latency can be viewed as being censored at each checkpoint $t$ because the latency values above $t$ are not revealed; alternatively, we could cast the problem as estimating whether a task will survive beyond the designated straggling latency. While the technical details of all three models differ greatly, they share a common assumption: that the underlying task latency behavior is known a priori. Tobit and Grabit assume the latency follows a Gaussian distribution, and they predict censored values according to this assumption. CoxPH makes a more relaxed assumption: rather than assuming a particular distribution, it assumes that all task's transformed survival curves (survival probability over time) have the same shape \cite{Hosmer2008-rr}, an assumption that does not hold in the online straggler prediction problem as heterogeneous behavior (either in the tasks themselves or the machines executing those tasks) is a common cause of straggling. In addition, the CoxPH model assumes that the relationship between a task's features and straggling behavior does not change over time, but this is not true in practice and \SYSTEM{} explicitly accounts for this behavior (\cref{sec:update}).

\subsection{Summary}

To address all the issues from the existing approaches applied to online straggler prediction, we propose \SYSTEM{}, a negative-unlabeled learning approach that requires no labeled positive examples of stragglers or assumptions on the latency distributions. \SYSTEM{} incorporates a data-driven way to learn the weighting function from feature space that can easily adapt to different types of distributions. Specifically, \SYSTEM{} uses the fact that unlabeled samples satisfy $y_i \geq \tau_t^{\rm \fast}$, rather than relying on any assumptions about the latency distribution. The next section describes how \SYSTEM{} learns a weighting function based on task features to mitigate the bias of supervised learning on the labeled samples. 

% The latency distributions for running jobs are usually not well-understood or easily-specified; e.g., long-tailed or short tailed. As shown in our experimental results, existing methods applying survival regression methods (with either linear or non-linear predictors) fail to accurately predict stragglers either because they only use

% their heavy reliance on pre-specified distributions. By contrast, \SYSTEM{} incorporates a data-driven way to learn the latency distribution that can easily adapt to different types of distributions. In particular, \SYSTEM{} leverages the key insight underlying survival regression models (i.e. that we must incorporate the knowledge that unlabeled samples satisfy $y_i \geq \tau_t^{\rm \fast}$), but rather than explicitly specifying $\mathbb{P}(Y|X=x_{ti})$, \SYSTEM{} learns a model that predicts $\mathbb{P}(Y \geq \tau_t^{\rm \fast} | X = x_{ti})$ and then uses this learned model to mitigate the bias of supervised regression on the labeled samples.

\section{The proposed approach: \SYSTEM{} }\label{sec:framework}

\SYSTEM{} is a novel negative-unlabeled learning approach for online straggler prediction without positive examples at training or assumptions on latency distributions. Specifically, \SYSTEM{} trains a new predictor for each job, customizing to that job's unique properties. The key idea is to first train a latency predictor using only finished tasks (i.e.,  non-stragglers), and then reweight those latency predictions based on a function of dissimilarity between finished and running tasks from the feature space. Algorithm~\ref{alg:nurd} summarizes \SYSTEM{}. In particular, there are three key components:
\begin{enumerate}
    \item \textbf{Training with finished tasks (\cref{sec:train-finished}).}
    \item \textbf{Reweighting based on feature space (\cref{sec:reweight-feature}).}
    % \item \textbf{Reweighting based on estimated latency space (\cref{sec:reweight-latency}).}
    \item \textbf{Updating models online (\cref{sec:update}).}
\end{enumerate}
Next, we describe each in detail.

\subsection{Training with Finished Tasks}\label{sec:train-finished}
\SYSTEM{} starts by training a latency predictor using the labeled finished tasks (i.e., non-stragglers, or negative examples). While any regression model can be applied, \SYSTEM{} uses gradient boosting trees due to its high predictive power in many settings~\cite{chen2016xgboost}. At the $t$-th time checkpoint for the $i$-th task, \SYSTEM{} trains the regression model $h_t$ so that the predicted latency is $\hat y_{ti} = h_t(x_{ti})$.  However, since it only trains on negative examples, the predictions will be heavily biased towards finished tasks (i.e., non-stragglers). To reduce such bias, \SYSTEM{} reweights the predictions, leading to the next step.

\subsection{Reweighting Based on Feature Space.}\label{sec:reweight-feature}
To reduce the bias from training on finished tasks only, \SYSTEM{} reweights $\hat y_{ti}$. At the $t$-th time checkpoint for the $i$-th task, \SYSTEM{} uses a weighting function $w_{ti}\in (0,1]$ such that
\begin{align}\label{eq:reweight}
\hat{y}_{ti}^{\rm adj} = \frac{\hat y_{ti}}{w_{ti}},
\end{align}
where $\hat{y}_{ti}^{\rm adj}$ is the adjusted latency prediction for the $i$-th task. Intuitively, when a running task's features are similar to finished tasks (i.e., non-stragglers), we want $w_{ti}$ to be relatively large (close to 1) such that $\hat{y}_{ti}^{\rm adj}$ does not change much from $\hat{y}_{ti}$. When a running task's features are different from finished tasks, we want $w_{ti}$ to be relatively small (close to 0) such that $\hat{y}_{ti}^{\rm adj}$ will be enlarged and more likely to exceed the latency threshold and be classified as a straggler.

To find a weighting function that matches our intuition, \SYSTEM{} uses \textbf{propensity score (PS)}, which is defined as the conditional probability of assignment to a particular group given a set of observed features~\cite{rosenbaum1983central}. In our case, we denote $z_{ti}$ as PS; i.e., the conditional probability that a task belongs to the class of finished tasks given its features at time checkpoint $t$:
\begin{align}\label{eq:ps}
z_{ti} = \mathbb{P}(y_i \le \tau_t^{\rm \fast}|x_{ti}).
\end{align}
In practice, $z_{ti}$ is usually estimated using logistic regression~\cite{cepeda2003comparison} because we have two known classes at the $t$-th checkpoint: finished tasks and running tasks. When $z_{ti}$ has relatively higher probability value (close to 1), it indicates that the $i$-th task has higher chances that it will finish soon (i.e., a non-straggler), and thus using it to reweight $\hat{y}_{ti}$ will not cause a large change in the latency prediction. In contrast, when $z_{ti}$ has a relatively low probability value (close to 0), it indicates that the $i$-th task has a higher chance that it will keep running (i.e., straggle), and thus using it to reweight $\hat{y}_{ti}$ will dilate the latency prediction and make it more likely to exceed the latency threshold.

\begin{algorithm*}[tb]
	\small
	\caption{Online straggler prediction with \SYSTEM{}.}
	\label{alg:nurd}
	\textbf{Input:} $T$: number of time checkpoints; $F_0$: list of tasks finished at initial checkpoint $t=0$; $R_0$: list of tasks running at initial checkpoint $t=0$; 
	$\tau^{\rm \slow} > 0$: latency threshold; $\alpha> 0$: calibration parameter; $\epsilon>0$: minimum positive weight. \\
	\textbf{Output: $\hat S$} 
	\begin{algorithmic}[1]
		\small
		\STATE 	Initialize $X_{\mathrm{fin}}$ and $X_{\mathrm{run}}$ with features from tasks in $F_0$ and $R_0$, i.e., 
		$X_{\mathrm{fin}} = \{x_{0i}; i \in F_0\},  X_{\mathrm{run}} = \{x_{0i}; i \in R_0\}$. 
		\STATE Initialize $Y_{\mathrm{fin}}$ in $F_0$, i.e., $Y_{\mathrm{fin}} = \{y_{i}; i \in F_0\}$.
		\STATE Straggler set $\hat S\gets \emptyset$.
		\STATE Compute centroids of $X_{\mathrm{fin}}$ and the rest running tasks $X_{\mathrm{run}}$, denoted as $c_{\mathrm{fin}}$ and $c_{\mathrm{run}}$. \label{line:centroid}
		\STATE $\rho = \|c_{\mathrm{fin}}\|^2/\|c_{\mathrm{run}} - c_{\mathrm{fin}}\|^2$.  \COMMENT{Compute latency indicator}
		\STATE $\delta = \frac{1}{1+\rho} - \alpha$. \label{line:delta} \COMMENT{Compute calibration term}
		\FOR{each time checkpoint $t=1,\ldots, T$}
		\STATE $\Delta_t \gets \{i \in R_{t-1} ; y_i \le \tau_t^{\rm \fast}\}$. \COMMENT{Update tasks finished between $t-1$ and $t$-th checkpoint}
		\STATE $F_t \gets F_{t-1}\bigcup \Delta_t$, $R_t \gets R_{t-1}\setminus \Delta_t$.  \COMMENT{Update sets of finished and running tasks}
		\STATE $X_{\mathrm{fin}}\gets X_{\mathrm{fin}} \bigcup \{x_{ti} ; i \in \Delta_t\}$, $Y_{\mathrm{fin}}\gets Y_{\mathrm{fin}}\bigcup \{y_{i}; i \in \Delta_t\}$, and $X_{\mathrm{run}}\gets \{x_{ti} ; i \in R_t\}$.
		\STATE Update latency prediction model $h_t$ and PS estimation model $g_t$ using updated $X_{\mathrm{fin}}, Y_{\mathrm{fin}}$, and $X_{\mathrm{run}}$. 
		\FOR{each task $i \in R_t$ }
		\STATE Get initial latency prediction $\hat{y}_{ti}=h_t(x_{ti})$.
		\STATE Get PS estimation $z_{ti}=g_t(x_{ti})$.
		\STATE $w_{ti} = \max(\epsilon,\min(z_{ti} + \delta, 1))$    \COMMENT{Construct final weighting function}  
		\STATE Get adjusted latency prediction $\hat{y}_{ti}^{\rm adj} = \frac{\hat y_{ti}}{w_{ti}}$.
		\IF {$\hat{y}_{ti}^{\rm adj} \geq \tau^{\rm \slow}$}
		\STATE $\hat S \gets \hat S \bigcup \{i\}$, $R_t \gets R_t \setminus \{i\}$  \COMMENT{Terminate the task $i$ if a straggler is predicted}
		\ENDIF
		\ENDFOR
		\ENDFOR
		\STATE \textbf{return} $\hat S$ 
	\end{algorithmic}
\end{algorithm*}

\begin{algorithm}[tb]
	\small
	\caption{Scheduling with more machines than tasks.}
	\label{alg:unlimited}
	\textbf{Input:} $T$: number of time checkpoints; $[n]$: set of running tasks
	\begin{algorithmic}[1]
		\small
		\FOR{each time checkpoint $t\in [T]$}
		\FOR{each running task $i\in [n]$}
		\IF{task $i$ is predicted to be a straggler}
		\STATE Terminate $i$ and relaunch it on a new machine.
		\STATE Update set of running tasks $[n] \gets [n] \setminus \{i\}$.
		\ELSE
		\STATE Go to next task.
		\ENDIF
		\ENDFOR
		\ENDFOR  
	\end{algorithmic}
\end{algorithm}	

\begin{algorithm}[tb]
	\small
	\caption{Scheduling with fewer machines than tasks.}
	\label{alg:limited}
	\textbf{Input:} $T$: number of time checkpoints; $[n]$: set of running tasks; $[m]$: set of available machines.
	\begin{algorithmic}[1]
		\small
		\FOR{each time checkpoint $t\in [T]$}
		\IF{new machine $k$ available}
		\STATE Update set of available machines $[m] \gets [m] \bigcup \{k\}$.
		\ENDIF
		\FOR{each running task $i\in [n]$}
		\IF{task $i$ is predicted to be a straggler}
		\IF{machines are available $[m] \neq \emptyset$ }
		\STATE Terminate $i$ and relaunch it on a new machine $j$.
		\STATE Update set of running tasks $[n] \gets [n] \setminus \{i\}$.
		\STATE Update set of available machines $[m] \gets [m] \setminus \{j\}$.
		\ELSE
		\STATE Go to next task.
		\ENDIF
		\ENDIF
		\ENDFOR
		\ENDFOR 
	\end{algorithmic}
\end{algorithm}	
  
% \subsection{Reweighting Based on Estimated Latency Space}\label{sec:reweight-latency}

\textbf{Calibration.} Different jobs have different latency distributions, and thus have different target latency thresholds to determine stragglers. To account for such difference, \SYSTEM{} adds a calibration term $\delta$ to the propensity scores to construct the final weighting function and balance the tradeoffs between true and false positives. Specifically, this calibration term is a function of the latency threshold; i.e., whether the latency threshold (e.g. p90) is greater than the half of the maximum latency.
\begin{figure}[h]
	\centering
	\begin{subfloat}
		\centering
		\includegraphics[width=0.48\linewidth]{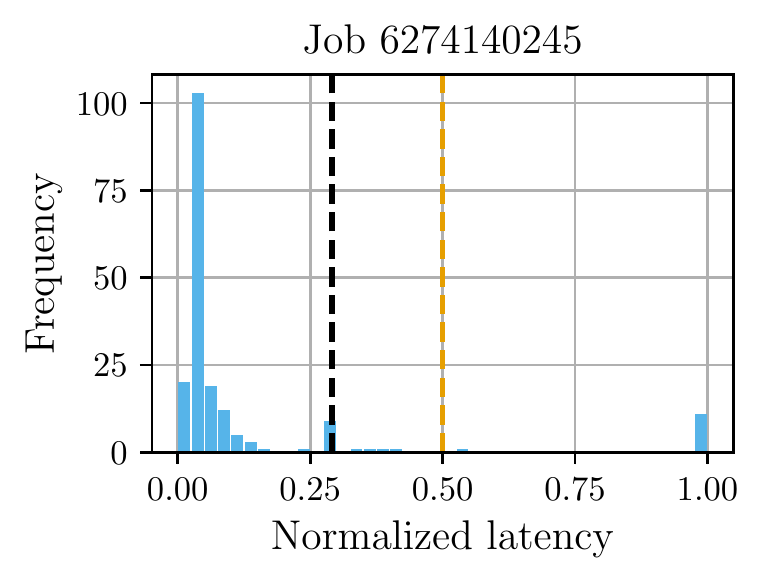}\vspace{-0.1in}
	\end{subfloat}
	\begin{subfloat}
		\centering
		\includegraphics[width=0.48\linewidth]{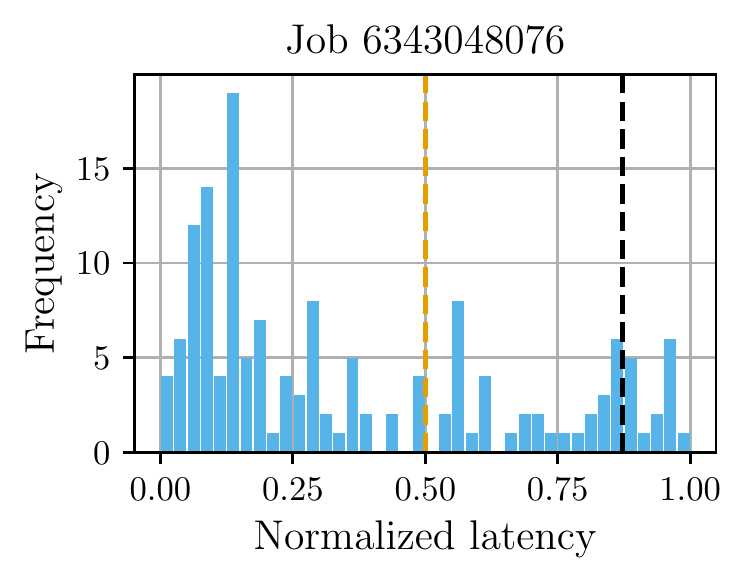}\vspace{-0.1in}
	\end{subfloat}
	\caption{Latency distributions for two Google jobs. The vertical dashed yellow line is half of the maximum normalized latency 0.5, and the dashed black line is the latency threshold (e.g. p90). }
	\label{fig:calibrate}
\end{figure}
\begin{itemize}
\item If the latency threshold is less than the half of the maximum latency (left side of Figure~\ref{fig:calibrate}), the latency threshold is relatively small compared to the maximum latency. To reduce the false positives in predictions, we hope to increase the weighting value such that $\hat{y}_{ti}$ will not be enlarged too much. Therefore, $\delta$ should be relatively large but not exceed 1;  i.e., $w_{ti} = \min(z_{ti}+\delta, 1)$.
\item If the latency threshold is greater than the half of the maximum latency (right side of Figure~\ref{fig:calibrate}), the latency threshold is relatively large. To reduce the false negatives in predictions, we hope to decrease the weighting value such that $\hat{y}_{ti}$ will be enlarged enough. Therefore, $\delta$ should be relatively small but not make $w_{ti}$ negative;  i.e., $w_{ti} = \max(\epsilon,\min(z_{ti} + \delta, 1))$, where $\epsilon$ is a small positive scalar.
\end{itemize}

With this insight, we address the remaining questions as follows: (1) how to determine whether the latency threshold is relatively large or small when the job is still running (since the actual task latencies are unknown), and (2) how to set $\delta$. 

\textbf{Determining whether latency threshold is relatively large or small.} As noted in \cref{sec:related} prior work for outlier detection and censored regression assumes a distribution (almost always Gaussian) for the values (in this case latency) they are trying to predict. Such an assumption makes determining the relative size of the latency threshold trivial.  A key distinction of \SYSTEM{} is that it makes no such assumption (and Figure \ref{fig:calibrate} shows that no single assumption would suffice).  Specifically, \SYSTEM{} estimates the relative magnitude of the latency threshold from tasks' observed features, rather than from assumptions about the latency distribution. Inspired by the insight that running tasks' features are different from those of finished tasks, \SYSTEM{} compares the feature centroids of finished tasks (non-stragglers) $c_{\mathrm{fin}}$ and those of running tasks $c_{\mathrm{run}}$ before starting prediction. Empirically, \SYSTEM{} computes $\rho = \|c_{\mathrm{fin}}\|^2/\|c_{\mathrm{run}} - c_{\mathrm{fin}}\|^2$. 

The intuition is that $\rho$ indicates how far potential stragglers are from non-stragglers. If potential stragglers are far from non-stragglers (left side of Figure \ref{fig:calibrate}), $c_{\mathrm{run}}$ is likely to be far from $c_{\mathrm{fin}}$ and $\rho\leq 1$. In this case, the unweighted latency prediction $\hat{y}_{ti}$ is easily pushed over the threshold, which makes PS overcorrect the latency predictions. Therefore, increasing the weight by making $\delta$ relatively large has little effect on true positives, but decreases false positives. In contrast, if potential stragglers are close to non-stragglers (right side of Figure \ref{fig:calibrate}), $c_{\mathrm{run}}$ is likely to be close to $c_{\mathrm{fin}}$ and $\rho > 1$. In this case, true stragglers may not be easily pushed over the threshold by $\hat{y}_{ti}$ because the features of all tasks are not so different, which makes PS alone not enough for reweighting. Therefore, reducing the weight by making $\delta$ relatively small increases the true positives significantly, despite a possible small increase in false positives.

\textbf{Setting $\delta$.} Assuming $\delta\in (-\alpha, \alpha)$, where $\alpha>0$. As discussed above, $\delta$ is a function of $\rho$. When the latency threshold is relatively small ($\rho\leq 1$), $\delta$ is relatively large. When the latency threshold is relatively large ($\rho> 1$), $\delta$ is relatively small. Therefore, \SYSTEM{} uses a function $f: \mathbb{R}_+ \times \mathbb{R} \to (-\alpha, \alpha)$ such that $\delta = f(\rho,\alpha)$:
\begin{align}
\delta = \frac{1}{1+\rho} - \alpha.
\end{align}

With $z_{ti}$, $\delta$, and $\hat{y}_{ti}$, \SYSTEM{} obtains 
\begin{align}
\hat{y}_{ti}^{\rm adj} = \frac{\hat{y}_{ti}}{\max(\epsilon,\min(z_{ti} + \delta, 1))}    
\end{align}
% $\hat{y}_{ti}^{\rm adj}$ based on Equation~\ref{eq:reweight}. 
\SYSTEM{} identifies stragglers if $\hat{y}_{ti}^{\rm adj} \ge \tau^{\rm \slow}$; i.e., the predicted latency exceeds the latency threshold. Since the exact value of $\tau^{\rm \slow}$ is unknown a priori, it can either be selected manually by users or automatically by techniques such as those in LinnOS~\cite{hao2020linnos} that estimate the inflection point in the latency CDF. Determining the latency threshold is beyond the scope of this work. Tests with a wide variety of thresholds show that \SYSTEM{} produces results that are robust to the different latency thresholds.

\subsection{Updating Models Online}\label{sec:update}

As the job is running, \SYSTEM{} accumulates examples of finished tasks at each checkpoint and \SYSTEM{} uses these new examples to update both the latency predictor $h_t$ and propensity score model in Equation~\ref{eq:ps} once the true task latencies are known. Thus, \SYSTEM{} improves prediction results as it collects more finished tasks.

\section{Scheduling}\label{sec:scheduler}

After \SYSTEM{} predicts a straggler, it can trigger the schedulers to mitigate straggling behavior, e.g., relaunching the predicted stragglers on other machines. To demonstrate how \SYSTEM{} can be used to reduce job completion time, we design schedulers to reassign tasks once a task is predicted to straggle. Our schedulers are based on the common strategy of relaunching predicted straggling tasks on new machines since it has been proved to be effective at mitigating stragglers~\cite{ananthanarayanan2013effective,lee2015outatime,ren2015hopper}. We consider two different situations: more machines than tasks and fewer machines than tasks.  
\begin{itemize}
    \item \textbf{More machines than tasks.} When more machines are available than tasks, a task that is predicted to be a straggler can be terminated and reassigned to a new machine immediately. Algorithm~\ref{alg:unlimited} summarizes the scheduling procedure when more machines are available than tasks.
    \item \textbf{Fewer machines than tasks.} When fewer machines are available than tasks, it is possible that not all predicted stragglers can be reassigned to new machines immediately. The scheduler needs to regularly check if there are new machines that just finished running tasks at each checkpoint. Should that be the case, these machines will also be considered for future assignment. Then, \SYSTEM{} will evaluate each running task and predicts if it will straggle. If that is the case and there are machines available, this task will be terminated and relaunched on a new machine immediately. Otherwise, the scheduler will move on to the next active task and wait for new machines at the next time checkpoint. Algorithm~\ref{alg:limited} summarizes the scheduling procedure when fewer machines are available than tasks.
\end{itemize}

\section{Experimental setup}\label{sec:setup}

\paragraph{Evaluation methodology.} We evaluate \SYSTEM{}'s ability to predict stragglers as jobs are running. We construct a simulator by parsing publicly available data traces and converting them into a time-series format; i.e., a series of the statistics available for each timestamp. The simulator replicates real execution by sending \SYSTEM{} the features that would be available at each time checkpoint. We use two public production traces from Google~\cite{reiss2011google} and Alibaba~\cite{alibaba} to demonstrate generality:
\begin{itemize}
\item \textbf{Google traces.} The Google traces include 29 days of data from 12.5K machines~\cite{reiss2011google,google-trace}. The trace consists of a number of jobs, each of which has tasks from 100 to 9999. We filter to only include production jobs with 100 or more tasks, which reduces the 650K jobs and 25M tasks to 8425 jobs and 1.1M tasks. There are 15 features per task including resource usage, microarchitectural, and scheduling behavior, shown in Table~\ref{tbl:google-feature}. 
%The trace contains time stamped data, which we arrange in time-order to form the time-series data set for our simulation framework.
\item \textbf{Alibaba traces.} The Alibaba traces include two subsets of traces~\cite{alibaba,alibaba-trace}: (1) 2017 traces consisting of 1.3K machines over 12 hours; (2) 2018 traces consisting of 4K machines over 8 days. The trace consists of a number of tasks, each of which has numerous instances. We filter the tasks to those with at least 100 instances, reducing to 1M tasks. There are 4 features per instance including CPU and memory usages, shown in Table~\ref{tbl:ab-feature}.
\end{itemize}

\begin{table}[!htb]
	\caption{Task features used in the Google Traces.}
	\label{tbl:google-feature}
	\footnotesize
	\begin{center}
		\begin{tabular}{l|l}
			\toprule
			\textbf{Feature} & \textbf{Description} \\	\midrule
			MCU      & Mean CPU usage  \\
			MAXCPU   & Maximum CPU usage \\
			SCPU     & Sampled CPU usage \\			
			CMU      & Canonical memory usage \\
			AMU      & Assigned memory usage \\
			MAXMU    & Maximum memory usage \\
			UPC      & Unmapped page cache memory usage \\
			TPC      & Total page cache memory usage \\
			MIO      & Mean disk I/O time \\
			MAXIO    & Maximum disk I/O time \\
			MDK      & Mean local disk space used \\
			CPI      &   Cycles per instruction  \\
			MAI      &   Memory accesses per instruction  \\
			EV      &    Number of times task is evicted  \\
			FL      &    Number of times task fails  \\
			\bottomrule
		\end{tabular}
	\end{center} 
	\vspace{-0.2in}
\end{table}

\begin{table}[!htb]
	\caption{Instance features used in the Alibaba Traces.}
	\label{tbl:ab-feature}
	\footnotesize
	\begin{center}
		\begin{tabular}{l|l}
			\toprule
			\textbf{Feature} & \textbf{Description}\\
			\midrule
			cpu$\_$avg  & Avg. CPU numbers of instance running  \\
			cpu$\_$max  & Max. CPU numbers of instance running \\
			mem$\_$avg  & Avg. normalized memory of instance running \\			
			mem$\_$max & Max. normalized memory of instance running \\
			\bottomrule
		\end{tabular}
	\end{center} 
	\vspace{-0.2in}
\end{table}

For all evaluations, \SYSTEM{} works on live data and makes predictions about which tasks will straggle without seeing any stragglers at training. The evaluations are run on a dual socket server with two 32-core Intel Xeon Gold 6242 processors, 192 GB RAM, and 2.80GHz clock speed. Several parameters are set as follows:

\textbf{Initial training data.} For each job, we first wait for 4\% of the entire tasks to complete as the initial training set, which are all non-stragglers. As the job is running, the training size increases as more tasks finish and are added to the training set. We only wait for a small amount of tasks to finish because we aim to mimic the real online experiments and start predicting as early as possible.

\textbf{Latency threshold.} We tested latency thresholds from p70 to p95 and the p90 results are representative of the average behavior over all those data points. We present results that use p90 as the latency threshold, i.e., any task's latency higher than 90th percentile latency is considered a straggler. 

\paragraph{Comparisons.} We compare to the following approaches:
\begin{itemize}
    \item \textbf{Supervised learning:} we compare to gradient boosting trees (GBTR), a widely-used regression model that achieves high predictive power in various prediction tasks~\cite{chen2016xgboost}. 
    \item \textbf{Outlier detection:} we compare to fourteen existing outlier detection methods with implementations available including ABOD~\cite{kriegel2008angle}, CBLOF~\cite{he2003discovering}, HBOS~\cite{goldstein2012histogram}, IFOREST~\cite{liu2008isolation}, KNN~\cite{ramaswamy2000efficient}, LOF~\cite{breunig2000lof}, MCD~\cite{hardin2004outlier}, OCSVM~\cite{scholkopf2001estimating}, PCA~\cite{shyu2003novel}, SOS~\cite{janssens2012stochastic}, LSCP~\cite{zhao2019lscp}, COF~\cite{tang2002enhancing}, SOD~\cite{kriegel2009outlier}, and XGBOD~\cite{zhao2018xgbod}, for which we use implementations from a state-of-the-art outlier detection library \texttt{PyOD}~\cite{zhao2019pyod}~\footnote{\url{https://github.com/yzhao062/pyod}}.
    \item \textbf{PU learning:} we compare to two PU learning methods with implementations available including PU-EN~\cite{elkan2008learning} and PU-BG~\cite{mordelet2014bagging}, for which we use implementations from \texttt{pulearn} package~\footnote{\url{https://pulearn.github.io/pulearn/}}.
    \item \textbf{Censored and survival regression:} we compare to three censored and survival regression methods with implementations available including Tobit~\cite{tobin1958estimation}, Grabit~\cite{sigrist2019grabit}, and Cox proportional hazard model~\cite{tian2005cox}, for which we use implementation from the author~\footnote{\url{https://github.com/fabsig/KTBoost}} and \texttt{lifelines} library~\footnote{\url{https://github.com/CamDavidsonPilon/lifelines/}}.
    \item \textbf{Wrangler:} we compare to Wrangler~\cite{yadwadkar2014wrangler}, a systems solution for straggler prediction by oversampling stragglers to address the issue of training set imbalance. It uses linear support vector machines for interpretability. Since Wrangler assumes positive examples of stragglers at training, we randomly sample 2/3 non-stragglers and stragglers from each job as training to mimic the same situation in the original paper. 
    \item \textbf{\SYSTEM{}-NC:} we compare to \SYSTEM{}-NC, a variant of \SYSTEM{} that does not including reweighting based on latency space, i.e., $w_{ti} = z_{ti}$ in Algorithm~\ref{alg:nurd}. This comparison aims to demonstrate the significance of accounting for the differences in latency thresholds for different jobs.
\end{itemize}

\paragraph{Hyperparameter tuning.} Since different jobs have different optimal hyperparameter settings, it is challenging to tune hyperparameters for each job individually for each method. To do a fair comparison, we select 6 jobs from each dataset to be used for hyperparameter tuning.  For the Google traces, We choose the same 6 representative jobs analyzed by humans in prior work~\cite{ZhengL18} as they are known to have mixed causes for straggling behavior. For the Alibaba traces, we choose the first 6 tasks in the dataset. Then for each learning method evaluated, we manually tune these jobs to find the optimal hyperparameters and apply them to all jobs. For \SYSTEM{} in particular, we set $\alpha=0.5$ and $\epsilon=0.05$ in Algorithm~\ref{alg:nurd}.

\section{Experimental evaluation}\label{sec:evaluation}

\begin{table*}[!htb]
\caption{Averaged results over all jobs for Google (15-dimensional features) and Alibaba (4-dimensional features) trace datasets. Higher is better for TPR and F1. Lower is better for FPR and FNR. The best F1 is in bold. } 
\label{tbl:pred-sum}
\begin{center}
\small
\begin{tabular}{l|l|llll|llll}
\toprule
                                     &              & \multicolumn{4}{c}{Google} & \multicolumn{4}{c}{Alibaba} \\ \midrule
                                     &              & TPR   & FPR  & FNR  & F1   & TPR   & FPR   & FNR  & F1   \\ \midrule
Supervised                           & GBTR         & 0.46  & 0.01 & 0.54 & 0.57 & 0.16  & 0.01  & 0.84 & 0.27 \\ \midrule
\multirow{14}{*}{Outlier detection}  & ABOD         & 0.95  & 0.56 & 0.05 & 0.29 & 0.02  & 0.01  & 0.98 & 0.04 \\
                                     & CBLOF        & 0.99  & 0.69 & 0.01 & 0.24 & 0.69  & 0.50  & 0.31 & 0.33 \\
                                     & HBOS         & 0.99  & 0.68 & 0.01 & 0.24 & 0.56  & 0.39  & 0.44 & 0.32 \\
                                     & IFOREST      & 0.94  & 0.52 & 0.06 & 0.31 & 0.58  & 0.45  & 0.42 & 0.29 \\
                                     & KNN          & 0.97  & 0.49 & 0.03 & 0.32 & 0.57  & 0.42  & 0.43 & 0.29 \\
                                     & LOF          & 0.90  & 0.45 & 0.10 & 0.33 & 0.39  & 0.24  & 0.61 & 0.25 \\
                                     & MCD          & 0.99  & 0.52 & 0.01 & 0.31 & 0.75  & 0.45  & 0.25 & 0.42 \\
                                     & OCSVM        & 0.91  & 0.47 & 0.09 & 0.32 & 0.56  & 0.41  & 0.44 & 0.29 \\
                                     & PCA          & 0.61  & 0.27 & 0.39 & 0.26 & 0.03  & 0.02  & 0.97 & 0.05 \\
                                     & SOS          & 0.22  & 0.24 & 0.78 & 0.12 & 0.08  & 0.08  & 0.92 & 0.11 \\
                                     & LSCP         & 0.96  & 0.51 & 0.04 & 0.29 & 0.65  & 0.44  & 0.35 & 0.35 \\
                                     & COF          & 0.27  & 0.21 & 0.73 & 0.14 & 0.10  & 0.07  & 0.90 & 0.14 \\
                                     & SOD          & 0.25  & 0.08 & 0.75 & 0.19 & 0.18  & 0.28  & 0.82 & 0.18 \\
                                     & XGBOD        & 0.58  & 0.18 & 0.42 & 0.28 & 0.48  & 0.14  & 0.52 & 0.38 \\ \midrule
\multirow{2}{*}{Positive-unlabeled}  & PU-EN        & 0.99  & 0.67 & 0.01 & 0.27 & 0.72  & 0.12  & 0.28 & 0.54 \\
                                     & PU-BG        & 0.99  & 0.99 & 0.01 & 0.16 & 0.86  & 0.16  & 0.14 & 0.57 \\ \midrule
\multirow{3}{*}{\shortstack{Censored and \\ survival regression}} & Tobit        & 0.97  & 0.52 & 0.03 & 0.46 & 0.35  & 0.02  & 0.65 & 0.32 \\
                                     & Grabit       & 0.91  & 0.17 & 0.09 & 0.70 & 0.72  & 0.19  & 0.28 & 0.49 \\
                                     & CoxPH        & 0.87  & 0.28 & 0.13 & 0.61 & 0.45  & 0.10  & 0.55 & 0.45 \\ \midrule
Systems                              & Wrangler     & 0.95  & 0.42 & 0.05 & 0.46 & 0.83  & 0.46  & 0.17 & 0.38 \\ \midrule
\multirow{2}{*}{Ours}                & \SYSTEM{}-NC & 0.96  & 0.60 & 0.04 & 0.42 & 0.98  & 0.57  & 0.02 & 0.37 \\
                                     & \SYSTEM{}    & 0.95  & 0.11 & 0.05 & \textbf{0.81} & 0.87  & 0.13  & 0.13 & \textbf{0.59} \\
		\bottomrule
\end{tabular}
\end{center}
\vspace{-0.2in}
\end{table*}

\begin{figure*}[!htb]
	\centering
	\includegraphics[width=0.9\textwidth]{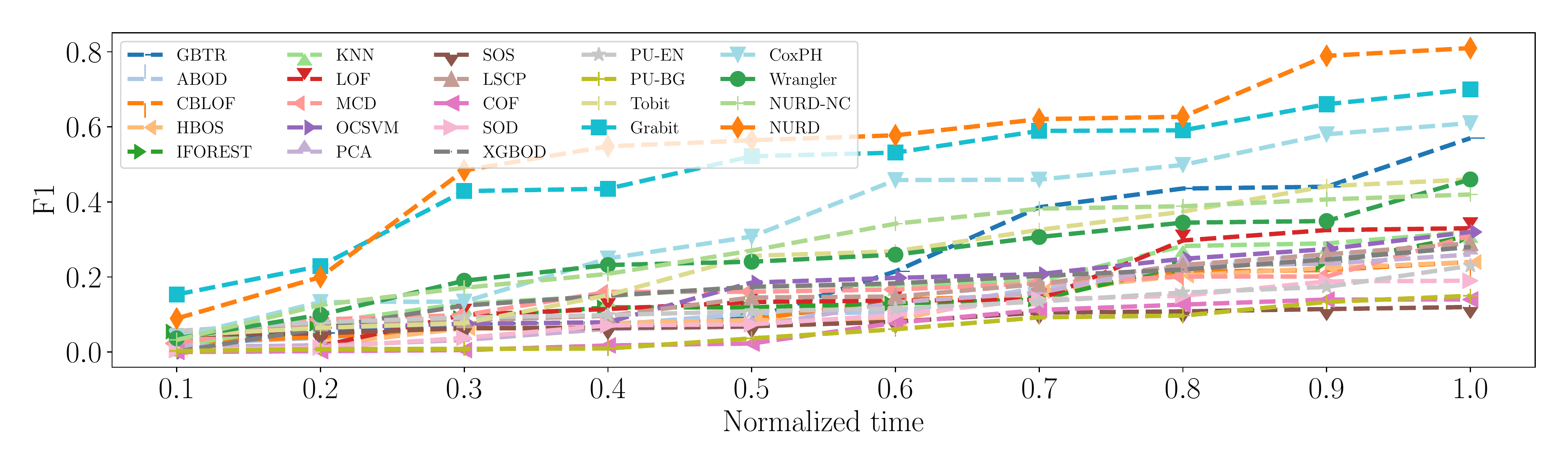}
	\vspace{-0.2in}
	\caption{F1 scores at different normalized time checkpoints for online straggler identification on Google traces (higher is better).} 
	\label{fig:google-early-curve}
	\vspace{-0.1in}
\end{figure*}

\begin{figure*}[!htb]
	\centering
	\includegraphics[width=0.9\textwidth]{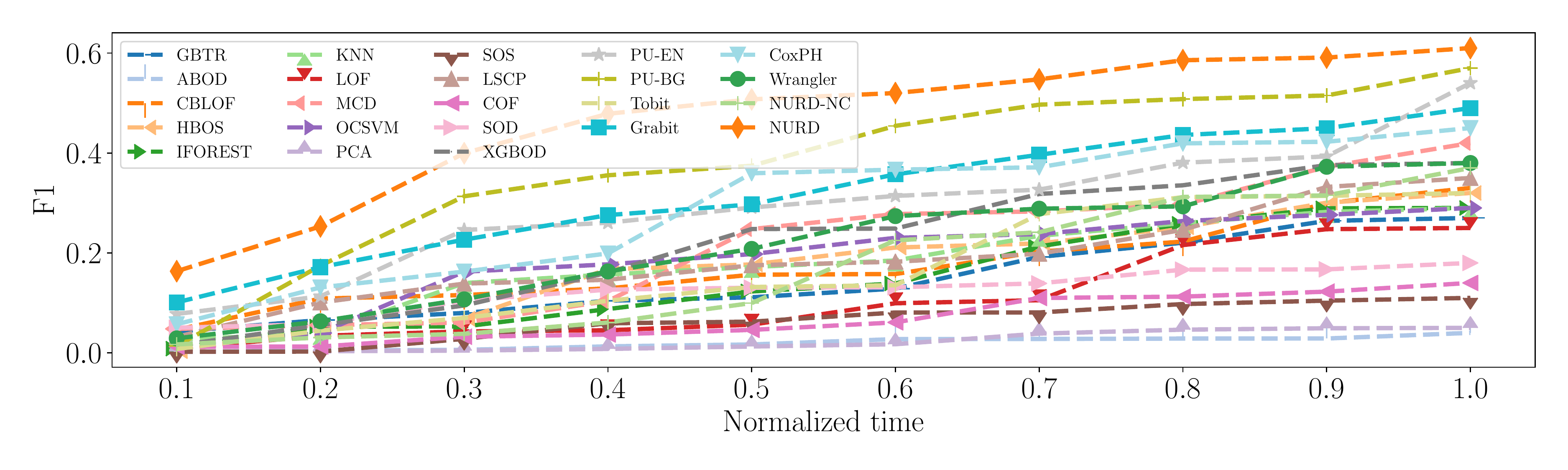}
	\vspace{-0.2in}
	\caption{F1 scores at different normalized time checkpoints for online straggler identification on Alibaba traces (higher is better).} 
	\label{fig:alibaba-early-curve}
	\vspace{-0.1in}
\end{figure*}

\begin{figure}[!htb]
	\centering
	\includegraphics[width=0.9\linewidth]{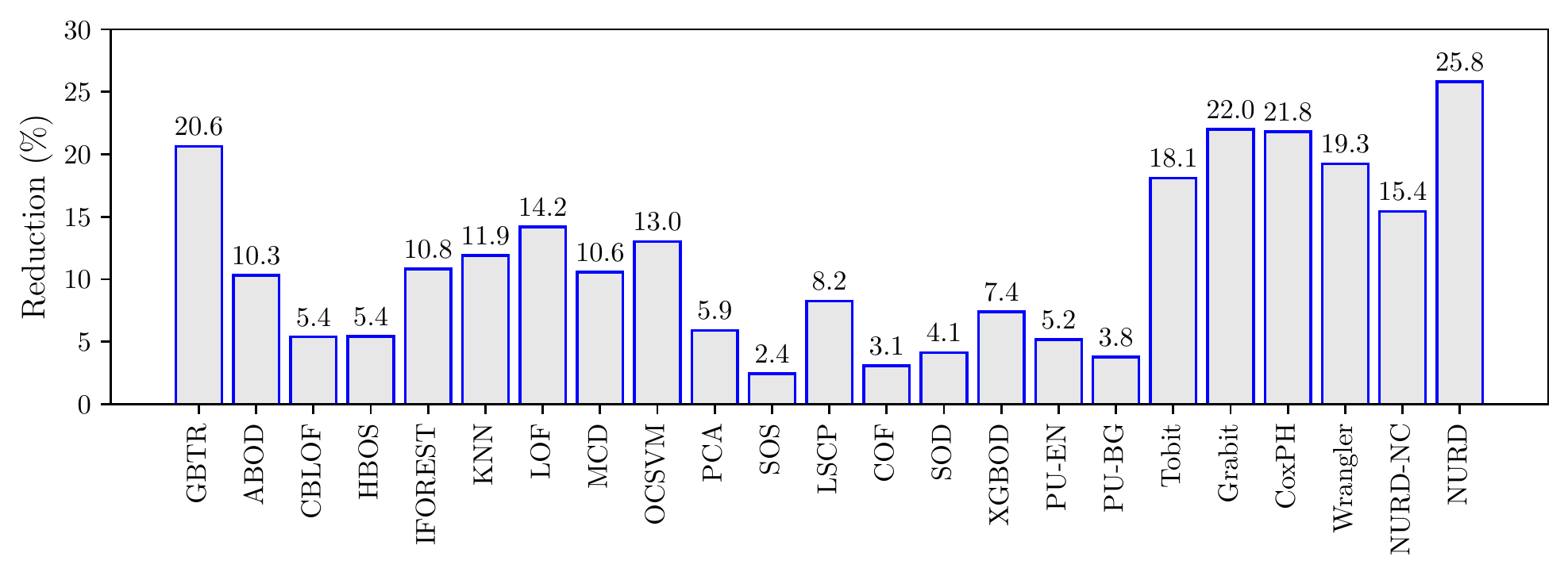}
	\vspace{-0.1in}
	\caption{Average reduction in job completion time with unlimited machines on Google traces (higher is better).} 
	\label{fig:google-unlimited}
	\vspace{-0.1in}
\end{figure}

\begin{figure}[!htb]
	\centering
	\includegraphics[width=0.9\linewidth]{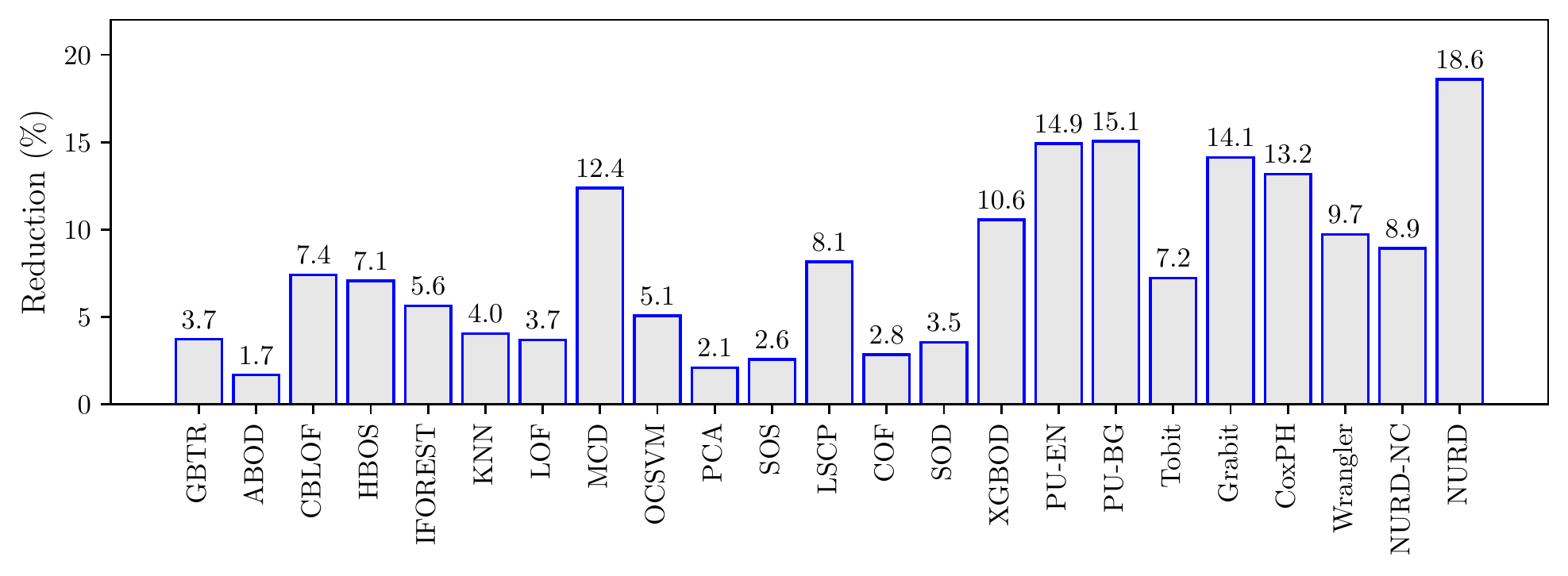}
	\vspace{-0.2in}
	\caption{Average reduction in job completion time with unlimited machines on Alibaba traces (higher is better).} 
	\label{fig:alibaba-unlimited}
	\vspace{-0.1in}
\end{figure}

\begin{figure*}[!htb]
	\centering
	\includegraphics[width=0.9\linewidth]{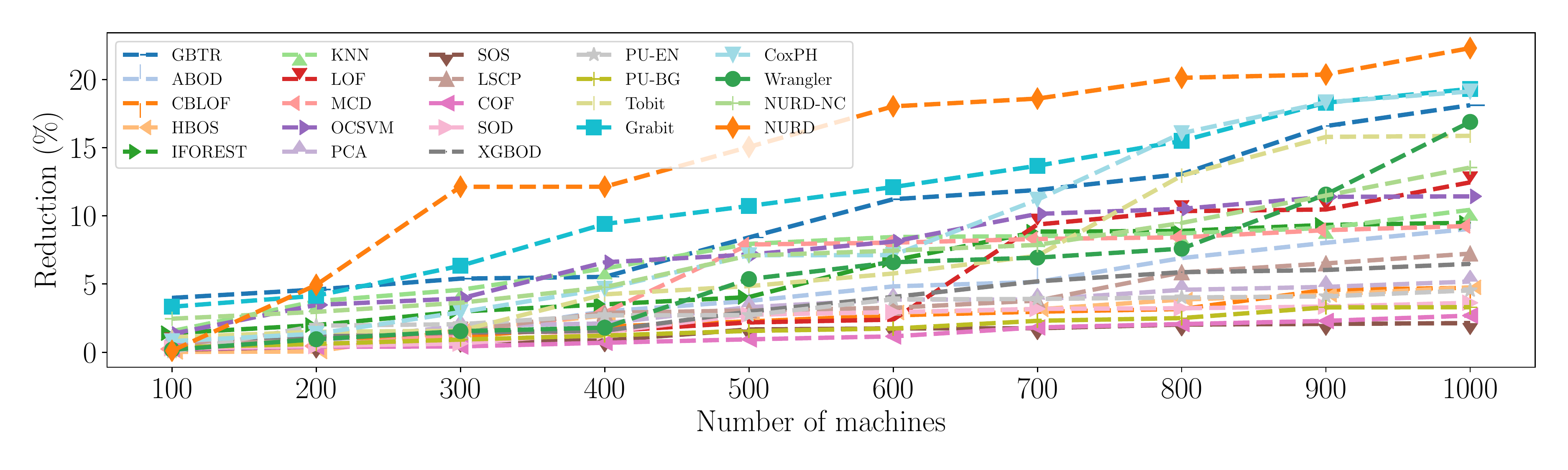}
	\vspace{-0.2in}
	\caption{Reduction in job completion time with different numbers of machines on Google traces (higher is better).} 
	\label{fig:google-limited}
	\vspace{-0.1in}
\end{figure*}

\begin{figure*}[!htb]
	\centering
	\includegraphics[width=0.9\linewidth]{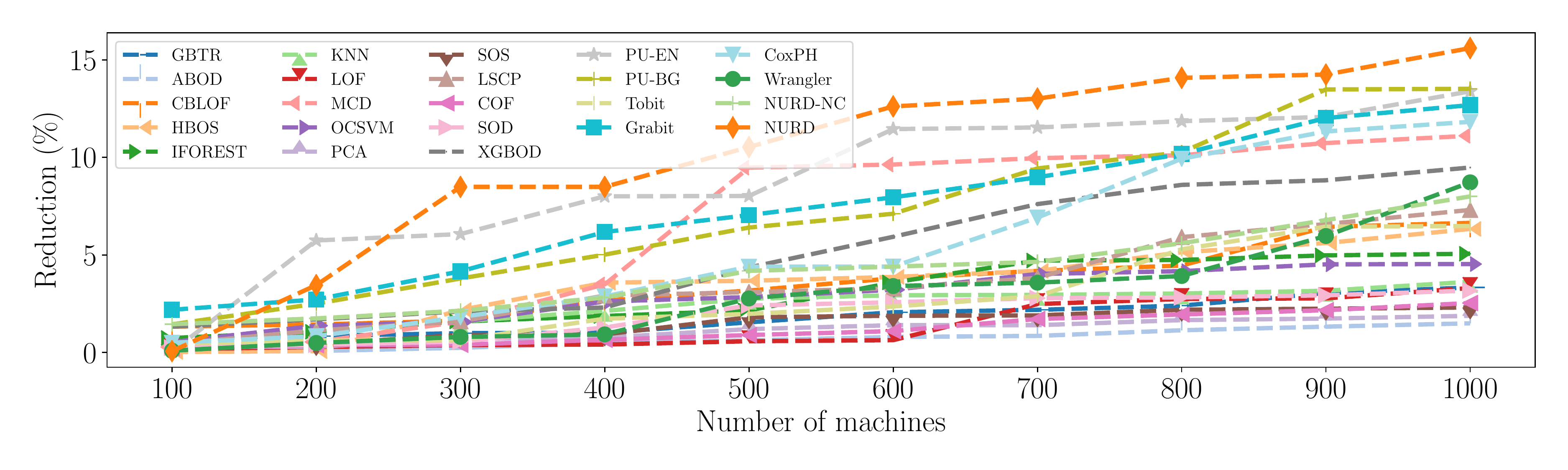}
	\vspace{-0.2in}
	\caption{Reduction in job completion time with different numbers of machines on Alibaba traces (higher is better).} 
	\label{fig:alibaba-limited}
	\vspace{-0.1in}
\end{figure*}

\begin{figure}[!htb]
	\centering
	\includegraphics[width=0.9\linewidth]{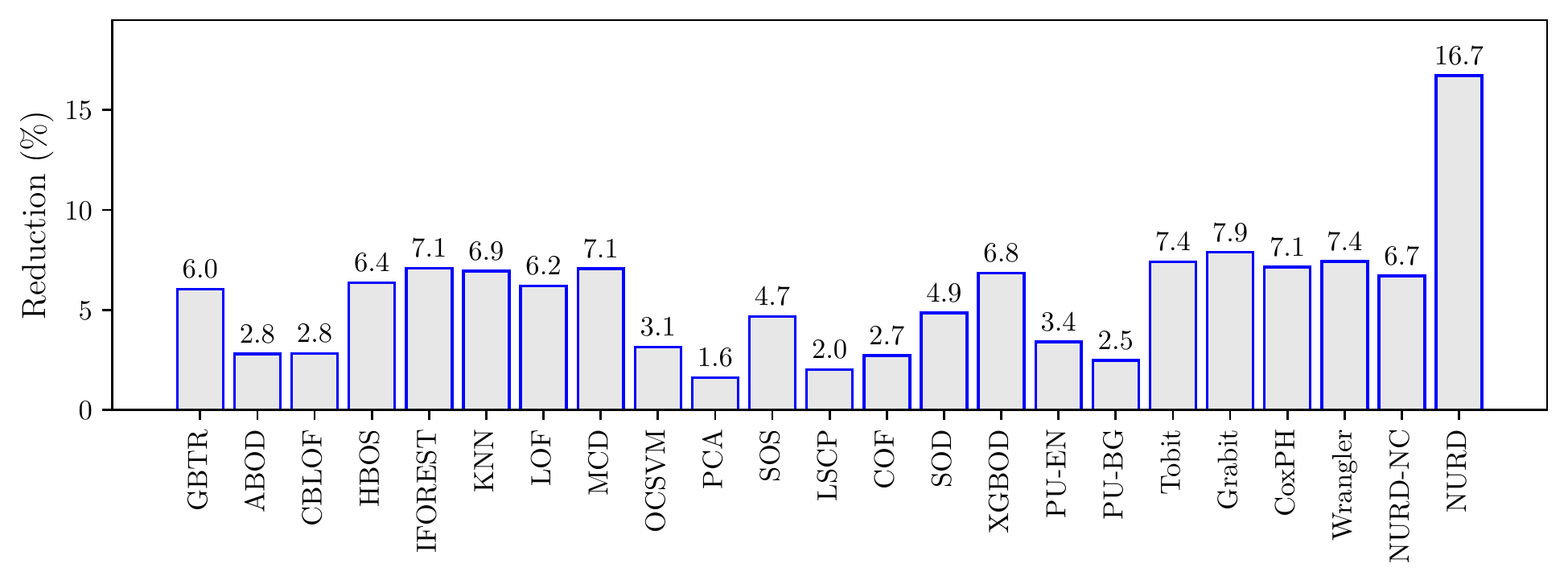}
	\vspace{-0.2in}
	\caption{Reduction in job completion time averaged over all number of machines on Google traces (higher is better).} 
	\label{fig:google-limited-bar}
	\vspace{-0.1in}
\end{figure}

\begin{figure}[!htb]
	\centering
	\includegraphics[width=0.9\linewidth]{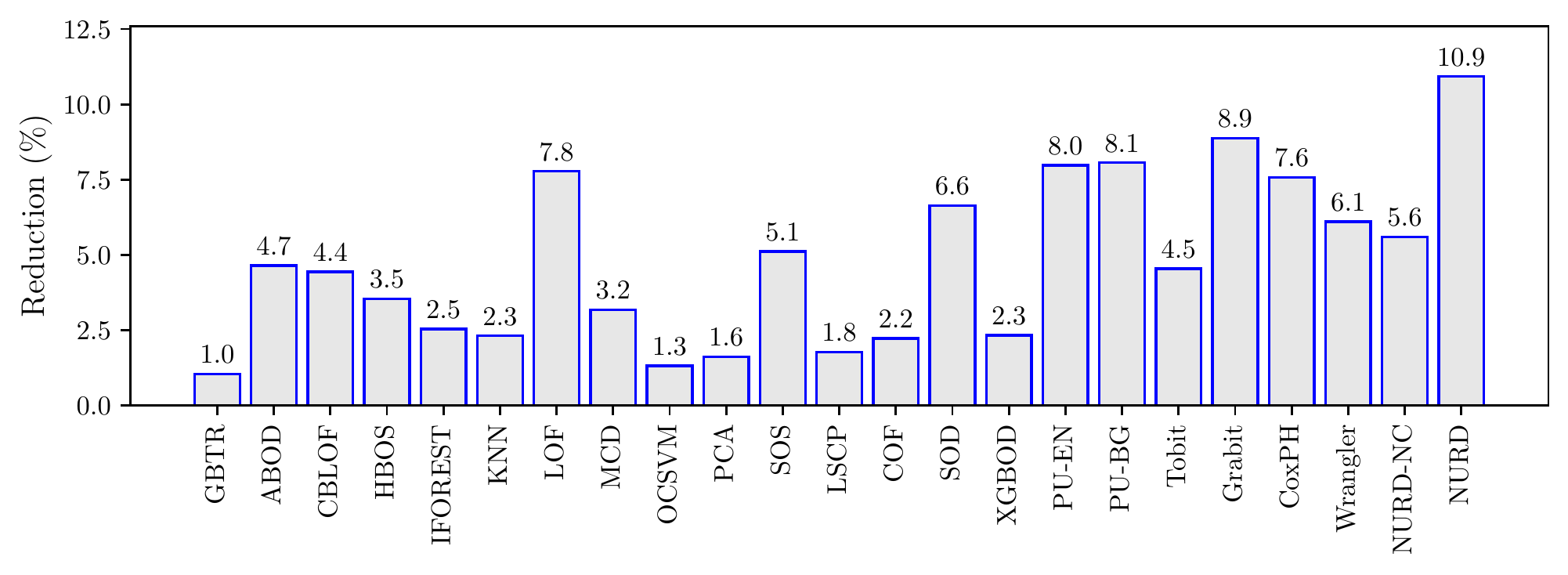}
	\vspace{-0.2in}
	\caption{Reduction in job completion time averaged over all number of machines on Alibaba traces (higher is better).} 
	\label{fig:alibaba-limited-bar}
	\vspace{-0.1in}
\end{figure}

% On two real-world production traces Google and Alibaba respectively, the key takeaways of our evaluation are: 
% \begin{itemize}
% \item \SYSTEM{} achieves at least 11 and 2 percentage points higher F1 scores than all other methods (Table~\ref{tbl:pred-sum}).
% \item \SYSTEM{} identifies more stragglers earlier than all other methods (Figure~\ref{fig:google-early-curve} and~\ref{fig:alibaba-early-curve}).
% \item \SYSTEM{} reduces at least 4 and 3.5 more percentage points completion time than all other methods when more machines are available than tasks (Figure~\ref{fig:google-unlimited} and~\ref{fig:alibaba-unlimited}).
% \item \SYSTEM{} reduces at least 4.7 and 8.8 more percentage points completion time than all other methods when fewer machines are available than tasks (Figure~\ref{fig:google-limited},~\ref{fig:alibaba-limited},~\ref{fig:google-limited-bar}, and~\ref{fig:alibaba-limited-bar}).
% \end{itemize}

The key takeaways of our evaluation are as follows. Compared to the best baseline approach:
\begin{itemize}
\item \SYSTEM{} achieves 11 and 2 percentage point increases in the F1 score for Google and Alibaba production traces, respectively (Table~\ref{tbl:pred-sum}).
\item \SYSTEM{} identifies stragglers earlier (Figure~\ref{fig:google-early-curve} and~\ref{fig:alibaba-early-curve}).
\item \SYSTEM{} has 3.8 and 3.5 percentage point improvements in job completion time when more machines are available than tasks (Figure~\ref{fig:google-unlimited} and~\ref{fig:alibaba-unlimited}).
\item \SYSTEM{} has 8.8 and 2.0 percentage point improvements in job completion time when fewer machines are available than tasks (Figure~\ref{fig:google-limited},~\ref{fig:alibaba-limited},~\ref{fig:google-limited-bar}, and~\ref{fig:alibaba-limited-bar}).
\end{itemize}

\subsection{How accurate are \SYSTEM{}'s predictions?}

Table~\ref{tbl:pred-sum} shows the average prediction results over all jobs. If a task is predicted to be a non-straggler at the $t$-th time point, it will be evaluated again at $(t+1)$-th time point if it is not finished. If a task is predicted to be a straggler, it will not be evaluated again. We use F1 score as the evaluation metric, and also show true positive rate (TPR), false positive rate (FPR), and false negative rate (FNR) to illustrate the tradeoffs between these metrics. 

We can see that the supervised learning method GBTR achieves low TPR and FPR because it is greatly impacted by a lack of stragglers at training: i.e., it predicts most tasks to be non-stragglers. The outlier detection methods have either both high TPR and FPR or both low TPR and FPR, which lead to low F1. It is not surprising since as unsupervised learning methods, the outlier detection methods do not utilize the knowledge of the observed latency. Therefore, it is difficult to assign an explicit separation boundary between stragglers and non-stragglers. As semi-supervised learning methods, PU-EN and PU-BG achieve impressive TPR, but fail to keep FPR consistently low. We notice that they tend to predict all tasks to be stragglers in early time checkpoints. Remember that there is a training and test distribution drift between training and test set for online straggler prediction. As classifiers rather than regressors, PU learners aggressively classify tasks that are different from training tasks (non-stragglers) to be stragglers. 

Censored and survival regression methods including Tobit, Grabit, and CoxPH are better than outlier detection and PU methods since they incorporate both features and latency at training. They are worse than \SYSTEM{} mainly due to the fact that they heavily rely on pre-specified distribution (e.g., Gaussian), while the latency distributions for different jobs are hard to specify in advance (e.g., some are long-tailed). Wrangler achieves both high TPR and FPR, mainly because its offline oversampling makes the prediction biased towards stragglers. Also, its linear classifier is limited in characterizing the nonlinear relationships between features and latency. Regarding our methods, both \SYSTEM{}-NC and \SYSTEM{} have high TPRs, but \SYSTEM{}-NC fails to keep FPR low while \SYSTEM{} does, which demonstrates the effectiveness of the calibration that accounts for the latency threshold in the weighting function. Overall, \SYSTEM{} has the best F1 scores: at least 11 and 2 percentage point increases relatively to the other methods for Google and Alibaba traces, respectively. 

Furthermore, we note that the best prior approaches are different on Google and Alibaba, while \SYSTEM{} achieves the best results on both datasets indicating its approach is more generalizable: Grabit's F1 score is second best (after \SYSTEM{} on Google, but 10 points worse than \SYSTEM{} on Alibaba, while PU-BG's F1 is second best for Alibaba, but 65 points worse on Google. These results demonstrate that \SYSTEM{}'s reweighting strategy has a dramatic positive effect on online straggler prediction because it makes no assumptions about the underlying data distributions or the existence of labels.

\subsection{Does \SYSTEM{} identify stragglers early?}

To illustrate the streaming results when the jobs are running, we compute F1 scores at different time checkpoints. Since different jobs have different total running time, we sample results from 10 time checkpoints for each job and regard them as normalized time. Figure~\ref{fig:google-early-curve} and~\ref{fig:alibaba-early-curve} show the results averaged over all jobs from Google and Alibaba traces, respectively, where the x-axis represents the normalized time between 0 and 1, and y-axis represents the F1 scores. We can see that, for Google traces, \SYSTEM{} outperforms all other methods at all time points except the very beginning. For Alibaba traces, \SYSTEM{} outperforms all other methods throughout the run time. These results show that \SYSTEM{} identifies more stragglers earlier than other methods.

\subsection{Does \SYSTEM{} improve completion time?}\label{sec:sched-results}

We evaluate how \SYSTEM{} contributes to reducing job completion time by augmenting existing schedulers with improved straggler predictions. The key idea of the schedulers described in~\cref{sec:scheduler} is to relaunch the task on a new machine once the task is predicted to straggle. In our experiments, the new completion time for a rescheduled task is randomly sampled from the existing execution times. We show results in the following two different situations.

\textbf{More machines than tasks.} When more machines are available than tasks for each job, a task that is predicted to straggle is terminated and relaunched on a new machine immediately (Algorithm~\ref{alg:unlimited}). Figure~\ref{fig:google-unlimited} and~\ref{fig:alibaba-unlimited} show the results for Google and Alibaba traces respectively, with the x-axis represents the method and the y-axis represents the reduction in job completion time (higher is better). We can see that \SYSTEM{} has the highest reductions, 25.8\% and 18.6\% for Google and Alibaba traces respectively, which are 3.8 and 3.5 percentage point improvements compared to the best baseline approach. \SYSTEM{} achieves these improvements due to its early and accurate predictions for stragglers. 

\textbf{Fewer machines than tasks.} When fewer machines are available than tasks for each job, the scheduler needs to check if a new machine is available for relaunch if a task is predicted to straggle (Algorithm~\ref{alg:limited}). We study how reduction in completion time will change as a function of the number of machines. Figure~\ref{fig:google-limited} and~\ref{fig:alibaba-limited} show the results, where the x-axis shows the number of machines from 100 to 900, and the y-axis shows the reduction in job completion time (higher is better). As the number of machines increases, the reductions also increase, and \SYSTEM{} has the highest reductions compared to all other methods at all numbers except the small size (i.e., 100 and 200). We also compute the reduction averaged over all number of machines in Figure~\ref{fig:google-limited-bar} and~\ref{fig:alibaba-limited-bar}, where x-axis represents different methods, and y-axis represents the average reduction. We can see that \SYSTEM{} \SYSTEM{} has the highest reductions, 16.7\% and 10.9\% for Google and Alibaba traces respectively, which are 8.8 and 2.0 percentage point improvements compared to the best baseline approach. These results demonstrate that \SYSTEM{} can be easily incorporated with different types of schedulers and effectively reduce the job completion time.

\section{Conclusion} \label{sec:conclusion}
\setlength{\parskip}{3pt}

This paper introduces \SYSTEM{}, a novel negative-unlabeled learning approach for online straggler prediction that requires no labeled positive examples or assumptions on latency distributions. The key idea is to train a predictor using finished tasks of non-stragglers to predict latency for unlabeled running tasks, and then reweight each unlabeled task's prediction based on a weighting function of its feature space.  Extensive evaluation results on two real-world production traces demonstrates the effectiveness of \SYSTEM{} for online straggler prediction. 
%As for limitations, \SYSTEM{} trains a unique predictor for each job assuming that each job's characteristics are unique, which holds in practice.
Looking ahead, there is a possibility to apply transfer learning~\cite{pan2009survey} to incorporate knowledge from other jobs to improve predictions. We will incorporate transfer learning and deploy our methods in real-world datacenters for future datacenter-scale research.

\section*{Acknowledgements}

We are grateful to Alex Renda who read the early draft of this work and provided extremely valuable feedback.  Yi Ding is supported by the National Science Foundation under Grant 2030859 to the Computing Research Association for the CIFellows Project and a Meta Research Award. Rebecca Willett is supported by NSF grant DMS-2023109 and AFOSR FA9550-18-1-0166. Henry Hoffmann is supported by NSF (grants CCF-2119184, CNS-1956180, CNS-1952050, CCF-1823032, CNS-1764039), ARO (grant W911NF1920321), and a DOE Early Career Award (grant DESC0014195 0003). Any opinions, findings, and conclusions or recommendations expressed in this material are those of the authors and do not necessarily reflect the views of the funding agencies.

% \pagebreak


\bibliography{reference}
\bibliographystyle{mlsys2022}

%%%%%%%%%%%%%%%%%%%%%%%%%%%%%%%%%%%%%%%%%%%%%%%%%%%%%%%%%%%%%%%%%%%%%%%%%%%%%%%
%%%%%%%%%%%%%%%%%%%%%%%%%%%%%%%%%%%%%%%%%%%%%%%%%%%%%%%%%%%%%%%%%%%%%%%%%%%%%%%
% SUPPLEMENTAL CONTENT AS APPENDIX AFTER REFERENCES
%%%%%%%%%%%%%%%%%%%%%%%%%%%%%%%%%%%%%%%%%%%%%%%%%%%%%%%%%%%%%%%%%%%%%%%%%%%%%%%
%%%%%%%%%%%%%%%%%%%%%%%%%%%%%%%%%%%%%%%%%%%%%%%%%%%%%%%%%%%%%%%%%%%%%%%%%%%%%%%
%\appendix
%\section{Please add supplemental material as appendix here}
%%
%Put anything that you might normally include after the references as an appendix here, {\it not in a separate supplementary file}. Upload your final camera-ready as a single pdf, including all appendices.

%%%%%%%%%%%%%%%%%%%%%%%%%%%%%%%%%%%%%%%%%%%%%%%%%%%%%%%%%%%%%%%%%%%%%%%%%%%%%%%
%%%%%%%%%%%%%%%%%%%%%%%%%%%%%%%%%%%%%%%%%%%%%%%%%%%%%%%%%%%%%%%%%%%%%%%%%%%%%%%

\end{document}